\documentclass{article}
\usepackage{spconf,amsmath,epsfig}
\usepackage[preprint]{spconfa4}
\usepackage{graphicx}
\usepackage{amsmath}
\usepackage{amssymb}
\newcommand{\eat}[1]{}

\usepackage{balance}  
\usepackage{booktabs}
\usepackage{tabularx}
\usepackage{amsfonts}
\usepackage{caption}
\usepackage{multirow}
\usepackage{algorithm}
\usepackage{algorithmic}
\usepackage{siunitx}

\usepackage{xcolor}
\usepackage{cite}
\usepackage{bm}
\DeclareMathOperator*{\argmax}{arg\,max}

\usepackage{mathtools}
\usepackage{subfig}
\usepackage{multirow}
\usepackage{bm}
\DeclarePairedDelimiterX{\infdivx}[2]{(}{)}{%
  #1\;\delimsize\|\;#2%
}

\usepackage[export]{adjustbox}

\let\OLDthebibliography\thebibliography
\renewcommand\thebibliography[1]{
  \OLDthebibliography{#1}
  \setlength{\parskip}{0pt}
  \setlength{\itemsep}{0pt plus 0.3ex}
}

\begin{document}\sloppy

\def\x{{\mathbf x}}
\def\L{{\cal L}}

\title{Discovering Domain Disentanglement for Generalized Multi-source Domain Adaptation}
%
\name{Zixin Wang, Yadan Luo, Peng-Fei Zhang, Sen Wang, Zi Huang}
\address{ The University of Queensland, Brisbane 4067, Australia\\{\small zixin.wang1@uqconnect.edu.au}, {\small\{lyadanluol, mima.zpf\}@gmail.com}, {\small sen.wang@uq.edu.au}, {\small huang@itee.uq.edu.au}}

\maketitle

%
\begin{abstract}
A typical multi-source domain adaptation (MSDA) approach aims to transfer knowledge learned from a set of labeled source domains, to an unlabeled target domain. Nevertheless, prior works strictly assume that each source domain shares the identical group of classes with the target domain, which could hardly be guaranteed as the target label space is not observable. In this paper, we consider a more versatile setting of MSDA, namely Generalized Multi-source Domain Adaptation, wherein the source domains are partially overlapped, and the target domain is allowed to contain novel categories that are not presented in any source domains. This new setting is more elusive than any existing domain adaptation protocols due to the coexistence of the domain and category shifts across the source and target domains. To address this issue, we propose a variational domain disentanglement (VDD) framework, which decomposes the domain representations and semantic features for each instance by encouraging dimension-wise independence. To identify the target samples of unknown classes, we leverage online pseudo labeling, which assigns the pseudo-labels to unlabeled target data based on the confidence scores. Quantitative and qualitative experiments conducted on two benchmark datasets demonstrate the validity of the proposed framework.
\end{abstract}

\section{Introduction}
Due to the heavy dependence on voluminous data for training, the performance of supervised deep learning models will vastly degrade when the labeled data is scarce. To tackle this issue, a possible solution is to combine multiple labeled training datasets (i.e., source domains) to learn a domain-agnostic model for an unlabeled test set (i.e., target domain), which refers to Multi-source Domain Adaptation (MSDA). MSDA allows the source and target data coming from different data distributions, e.g., using images taken from foggy, rainy, and snowy days to classify the ones collected from sunny days. 
\begin{figure}[H]
\centering
\includegraphics[width=1\linewidth]{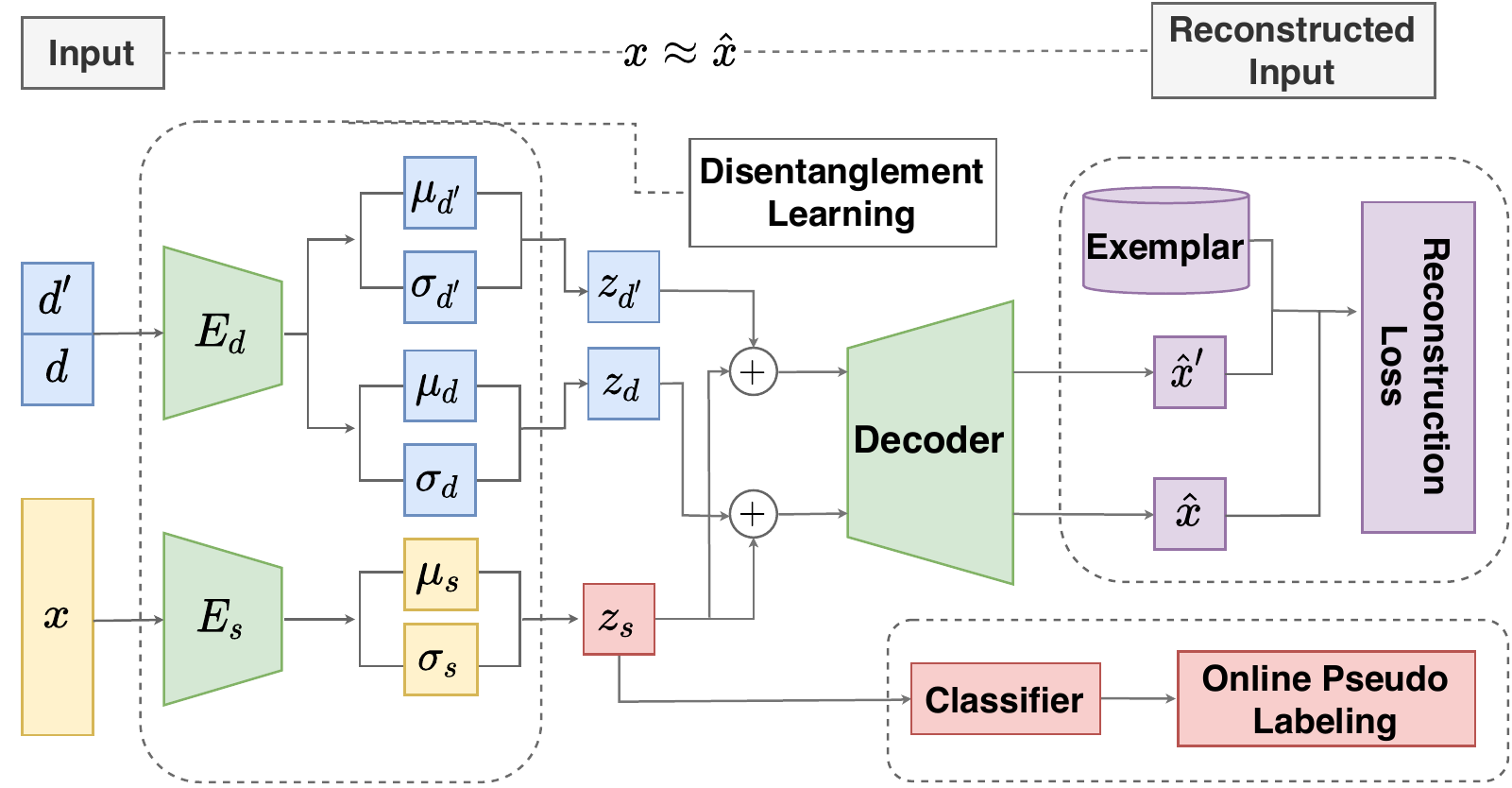}
\caption{Variational Domain Disentanglement (VDD).}\label{fig:model}
\end{figure}
However, it is challenging to directly apply MSDA into real-world scenarios due to its over-rigorous assumptions on the label space: 
(1) MSDA requires all source domains to have an identical label set, which could be easily violated in practice. It is highly likely that one source domain may only share \textit{a part of} the classes with another, while the rest of classes are privately preserved; \eat{For instance, one of the source domains contains images with the label car, book, and tree, while another source domain contains book, car, and pen. }
(2) MSDA is commonly under a closed-set setting where the source and target domains are expected to contain the same classes.  As the target labels are not available during training, it is more reasonable to consider an \textit{open-set} setting that enables the target domain to contain novel classes that are not presented in any source domains.

In this paper, we relax the label constraint of MSDA and propose a novel yet challenging setting, namely \textit{generalized multi-source domain adaptation} (GMDA). GMDA aims to learn a model from multiple complementary source domains and generalize to the unlabeled target domain with the presence of unknown classes. The core challenge of this task is simultaneously learning to reduce the influence of both category shift (i.e., partially shared source classes and the unknowns in target) and domain shift\eat{ \cite{DBLP:conf/aaai/SunFS16}}. Although distribution matching methods \cite{DBLP:conf/cvpr/ManciniPBC018,DBLP:conf/nips/LiMDC18} or adversarial learning approaches \cite{DBLP:conf/nips/ZhaoZWMCG18} can address the domain shift issue, these prior works could suffer the negative-transfer and class misalignment triggered by unshared categories residing in both the source and target domains. 

To address the above issue, we propose a \textit{Variational Domain Disentanglement (VDD)} framework to bridge both the category and domain gap hidden in visual recognition tasks. Specifically, we design a dual-branch (i.e., paralleled encoders) VAE framework to decompose the representation of each source and target instance into domain-specific and sample-specific features, where only the latter one is used for classification. As illustrated in Fig. \ref{fig:model}, the sample and domain encoders (i.e., $E_{s}$ and $E_{d}$) take the raw data and their paired domain labels as input to generate latent vectors, which are then concatenated and passed to the decoder for image reconstruction. We establish two strategies, exemplar learning, and disentanglement learning, to further boost domain disentanglement. For exemplar learning, the model learn to reconstruct the image to resemble an `exemplar' that has the same semantics but a different style as the raw data by replacing the ground-truth domain label of the original data with a `fake' one. Exemplars are randomly selected from each category and source domain, which helps to stabilize the image reconstruction and minimize the intra-domain variance.
Regarding the disentanglement learning, we substitute the KL divergence in the evidence lower bound (ELBO) of na\"ive VAE with total correlation (TC) to learn orthogonal domain representations to gain superior generalization capacity. The disentangled factor learning explicitly reinforces the domain-invariance of the learned sample embedding, which can be more correlated to the intrinsic semantics. In addition, we propose an online pseudo labeling regime to identify the most/least confident target data as the shared known/unknown class to supervise the model in turn. After that, the latent representations of images from the sample encoder can finally become domain-invariant, which then be fed into the classifier for logits. Empirical results show that VDD achieves state-of-the-art performance against existing single-source, multi-source, and open-set domain adaptation methods. Source code is released in the Github repo\footnote{https://github.com/Jo-wang/VDD}.

\section{Related Work}
Unsupervised domain adaptation (UDA) studies apply the model trained in a labeled source domain to an unlabeled target domain, where no label information is at one's disposal. Many challenging applications under different UDA settings  \cite{DBLP:conf/icml/LuoWHB20,DBLP:conf/mmasia/ChenLB21} were proposed. While more graph based methods can be found at \cite{DBLP:conf/mm/LuoHW0B20,DBLP:conf/mm/WangLHB20}. Multi-source Domain Adaptation (MSDA) typically assumes that the target distribution is a mixture of the source distributions, and search for the optimal combination of the source hypothesis \cite{DBLP:conf/nips/MansourMR08}. By considering the existence of category shift, Xu \textit{et al}. \cite{cocktail} proposed Deep Cocktail Network (DCTN), which leverages the K-way adversary to determine the similarity between the source and target domains and re-weights the respective domain-specific classifier. Another stream of work follows the spirit of learning domain-invariant features by adversarial learning \cite{DBLP:conf/nips/ZhaoZWMCG18} or distribution matching \cite{DBLP:conf/cvpr/ManciniPBC018,DBLP:conf/nips/LiMDC18,moment} rather than weighting different hypotheses to cope with the domain shift. Han \textit{et al}. \cite{DBLP:conf/nips/ZhaoZWMCG18} extended the domain adversarial neural networks to the multi-source setting. Li \textit{et al}. \cite{DBLP:conf/nips/LiMDC18} additionally considered the relationship between source pairs and derived a tighter bound on the weighted multi-source discrepancy. Similarly, Peng \textit{et al}. \cite{moment} proposed the MSDA mechanism aligned the moments of feature distributions between each source-source and source-target pair. Wang \textit{et al}. \cite{DBLP:conf/eccv/WangXN020} proposed Learning to Combing for Multi-Source Domain Adaptation (Ltc-MSDA), which leverages graph neural networks to propagate cross-domain information within the subgraph for each class. Yang \textit{et al}. \cite{DBLP:conf/eccv/YangBLS20} re-weighted the source samples by learning a dynamic curriculum, which progressively learns which samples are helpful for adaptation. While the notion of MSDA has been widely exploited in the past, existing algorithms are developed under the strong constraint of label sets for both source and target domains, i.e., assuming each domain has identical sets of classes. Recently, \eat{tackling domain adaptation issues in a } variational disentanglement provides new solutions to large amounts of deep vision tasks \cite{DBLP:conf/iccv/ChenLQ0H0021,DBLP:conf/nips/LiuLYW18}. For domain adaptation, Cao \textit{et al}. \cite{DBLP:journals/corr/abs-1805-08019} proposed frameworks with adversarial training to disentangle feature in the latent space. While TLR \cite{DBLP:conf/icmcs/XiaoDWZHL18} aims at learning a latent representation to solve the domain shift problem.

\section{Method}
\subsection{Problem Formulation and Notations}\label{sec:pf}
\textbf{Multi-source Domain Adaptation (MSDA)} aims to transfer a model trained on $N$ labeled source domains $\mathcal{D}_S = \left \{ D_{1}, D_{2}, \ldots D_{N} \right\} $ to an unlabeled target domain $\mathcal{D}_T$. Each source domain contains \textit{i.i.d.} sampled data $X_{S_{i}}= \left \{ x_{1}, \dots, x_{n} \right \}$ with the respective labels $Y_{S_i}=\{y_{1},\ldots$, $y_{n}\}\in\mathcal{Y}$, where $\mathcal{Y}$ is the label space. The ultimate target is to learn a domain-invariant model $f_{\theta}:\mathcal{X}_T\rightarrow \mathcal{Y}$ parameterized by $\theta$ that generalizes the target samples. 

\noindent \textbf{Generalized Multi-source Domain Adaptation (GMDA)}. Different from MSDA which assumes each domain share the same label space $\mathcal{Y}\in\{1,\ldots, C\}$, GMDA relaxes this constraint on both sides. Specifically, it allows the source domains to share a part of classes $Y_{ij} = Y_{S_i} \cap Y_{S_j}$ and have the label sets private to the other source domain $\bar{Y}_{ij} = Y_{S_i} \setminus Y_{ij}$. We indicate all source labels $Y_{S} = \bigcup Y_{S_{i}}$, and the target label set is a superset of $Y_S$, i.e., $Y_{T} = \{Y_{S}, C+1\}$, where $C+1$ is the unknown class. \eat{Note that the domain label is also available in this setting, where it is set to $D = \{1, 2, ... N\}$ for N domains.} 

\noindent \textbf{Variational Domain Disentanglement (VDD)}. The overview of the proposed VDD framework is presented in Fig. \ref{fig:model}. In each batch, we sample data from each source domain and target domain and denote them as $x$ for brevity. We use $d$ as the domain label of $x$, and randomly sample a fake domain label $d'$ that $d'\neq d$. The dual-branch encoder consists of a domain encoder $E_{d}(\cdot ; \phi_d)$ and a sample encoder $E_{s}(\cdot; \phi_s)$ that produces a variational probability model $q_{\phi}(z|x,d)$. Here the variational parameters are $\phi = \{\phi_d, \phi_s\}$. The latent variable for the domain label, fake domain label and sample are $z_d$, $z_{d'}$ and $z_s \in\mathbb{R}^{m}$. We concatenate the $z_d$, $z_{d'}$ with $z_s$ and feed them to the decoder network $G(\cdot; \theta)$\eat{, which is parameterized by $\theta$} to reconstruct sample $\hat{x} = G([z_s; z_d])$ and a fake sample $\hat{x}' = G([z_s; z_{d'}])$. On top of the encoder, a classifier $h(\cdot)$ is trained to take $z_s$ as input to make predictions. We will detail each component in the following sections.

\subsection{Dual-branch Variational Autoencoders}\label{sec:vdd}
The main challenge of the GMDA task is to learn a domain-agnostic model while avoiding negative transfer caused by the category shift and domain gap. The key idea is to separate the \textit{domain-specific} information and the \textit{domain-invariant} features that are only related to samples' semantics. VAEs \cite{DBLP:journals/corr/KingmaW13} are generative models that jointly train both probabilistic encoder and decoder, wherein the encoder learns to generate the latent variable $z$ following the pre-defined prior $p(z)$. In particular, to optimize the reconstructed output, we maximize the following evidence lower bound (ELBO),   
\begin{equation}\label{eq:orig_vae}
    \mathcal{L}_{\text{VAE}} = \mathbb{E}_{q_{\phi}(z|x)}[-\log(p_\theta(x|z))] + D_{\mathrm{KL}}(q_{\phi}(z|x) \| p(z)),
\end{equation}
where $z$ is the concatenation of the sampled vector $z_s$ and $z_d$, i.e., $z=[z_s;z_d]$. The first term can be interpreted as a reconstruction loss $\int q_\phi(z|x)\times \|x-G(z)\|^2 dz$, which aims to recover the original image with the domain vector $z_d$ and sample vector $z_s$. The second term calculates the KL-divergence, which penalizes the deviation of the latent feature $z$ from the prior distribution $p(z)$. Without loss of generality, we use a Gaussian distribution as a prior.

\subsection{Exemplar Learning} \label{exemplar}
In order to decouple the domain features from input image, we additionally generate `fake' reconstructed image $\hat{x}' = G(z')$ by feeding the concatenation of $z_{d'}$ and $z_{s}$, i.e., $z'=[z_s; z_{d'}]$ into the decoder. The generation of $\hat{x}'$ is supervised by a novel \textit{exemplar learning}:  for each category $c$ in each domain $i$, we randomly choose one sample as an exemplar $v_{i}^c$ and store it in the exemplar pool $\mathcal{V} = \{v_i^1, \ldots, v_i^C\}_{i=1}^{N}$. Then the corresponding exemplar is used as a ground-truth of the fake reconstructed image that shares the same class and domain label to calculate the reconstruction loss:
\begin{equation} \label{loss:diverse}
    \mathcal{L}_{exe} = \mathbb{E}_{q_{\phi}(z'|v)}[-\log(p_\theta(v|z'))] + D_{\mathrm{KL}}(q_{\phi}(z'|v) \| p(z')),
\end{equation}
where $v$ stands for the selected exemplar that matches the class of $x$ and the domain index of $d'$. The motivation of applying the exemplar learning is to (1) stabilize the image reconstruction, and the separation of domains and instances; (2) implicitly minimize the variance of the latent variable $z$ and force it to be more correlated to the semantics, hereby improving the classification accuracy.

\subsection{Disentangled Factor Learning} \label{disent}
 To achieve domain disentanglement and domain generalization with explainability, we employ \textit{disentangled factor learning}. The key insight is to make domain features orthogonal to each other in the latent space, i.e., $z_{d_{i}} \perp z_{d_{j}}$ where $z_{d_j}$ indicates the $j$-th dimension of the domain latent vector $z_d$. Each dimension of $z_{d}$ is independent, so that the representations can potentially generalize to other unseen domains. To facilitate this, we substitute the KL-divergence terms in Eq.\eqref{eq:orig_vae} and Eq.\eqref{loss:diverse}:
\begin{equation} \label{loss:vae}
D_{\mathrm{KL}}(q_{\phi}(z|x) \| p(z)) = I_{\phi_{d}}(z_{d} ; n)-\beta \mathcal{L}_{tc}-\xi\mathcal{L}_{dim},\\
\end{equation}
\begin{equation} \label{vae-detail}
 \begin{split}
&I_{\phi_{d}}(z_{d} ; n) = D_{\mathrm{KL}}(q_{\phi_{d}}(z_{d}, n) \| q_{\phi_{d}}(z_{d}) p(n)), \\
 &\mathcal{L}_{tc} = D_{\mathrm{KL}}(q_{\phi_d}(z_d) \| \prod_{j} q_{\phi_d}(z_{d_j})),\\
 &\mathcal{L}_{dim} = \sum_{j} D_{\mathrm{KL}}(q_{\phi_d}(z_{d_j})\|p(z_{d_j})),
    \end{split}
\end{equation}
where $n$ indicates the sample index of the $x$, and $p(n)$ is $\frac{1}{n}$. The index-code mutual information $I_{\phi_{d}}(z_{d} ; n)$ is calculated between data variable and latent variable on the basis of empirical data distribution $q_{\phi_{d}}(z_d, n)$. $\mathcal{L}_{tc}$ is the total correlation (TC) that measures the dependency between the variables so that reducing TC could benefit learning statistically independent factors in the data distribution. $\mathcal{L}_{dim}$ is the dimension-wise KL-divergence that limits individual latent dimensions to deviate too far from their respective priors. 

\subsection{Online Pseudo Labeling}\label{sec:online}
Aiming at improving the classification performance on the unlabeled target domain, we choose to perform batch-wise pseudo labeling, which helps propagate the knowledge from labeled source data to unlabeled target data. In particular, based on the prediction $\hat{p}(x)\in\mathbb{R}^{C+1}$ produced from the classifier for each target sample $x$ ($x\sim D_T$), we assign the pseudo label $\hat{y}$ by thresholding the confidence with $\delta_{known}$ and $\delta _{unk}$. To be detailed, 
\\
\begin{equation}
    \hat{y} =\begin{cases}
		    \argmax\hat{p}(x), & \text{if }~\max\hat{p}(x)> \delta_{known}\\
            C + 1, & \text{if }~\max\hat{p}(x)< \delta_{unk} \\
            \emptyset.   & \text{otherwise}
		 \end{cases}
\end{equation}

 After that, the pseudo labeled data is utilized to train the model in turn with the cross-entropy loss:
\begin{equation}
\mathcal{L}_{\mathrm{pseudo}}=\mathbb{E}_{x \sim D_{p}} -\hat{y}\log \hat{p}(x), 
\end{equation}
where $D_{p} = \{(x_t, \hat{y}) | \hat{y}\neq \emptyset\}$ is defined as the set of pseudo labeled target data and the respective pseudo labels. To further remain the inter- and inner-relationship between classes and avoid overfitting, the soft entropy $\mathcal{L}_t$ as applied as a regularizer on the target data:
\begin{equation}\label{eq:cls}
    \begin{split}
\mathcal{L}_{\mathrm{t}} = \mathbb{E}_{x\sim D_{T}} -\hat{p}(x) \log \hat{p}(x).
\end{split}
\end{equation}
While for the source data, cross-entropy loss $\mathcal{L}_{s}$ is employed:
\begin{equation}\label{eq:cls1}
    \begin{split}
         \mathcal{L}_{\mathrm{s}} = \mathbb{E}_{x\sim D_{S}} -y \log \hat{p}(x) .
    \end{split}
\end{equation}

\subsection{Joint Optimization}\label{sec:obj}
To conclude, the overall objective of the proposed model is to minimize:
\begin{eqnarray}
\mathcal{L} = \lambda\mathcal{L}_{s} + \mathcal{L}_{t} + \gamma\mathcal{L}_{\mathrm{VAE}} + \alpha \mathcal{L}_{exe} + \mathcal{L}_{pseudo}.
\end{eqnarray}

\section{Experiments}
\begin{figure}[t] 
    \centering
    \setlength{\abovecaptionskip}{0.cm}
    \subfloat[][CIFAR-10-Corrupted level 3]{\includegraphics[width=0.48\linewidth, height=4cm]{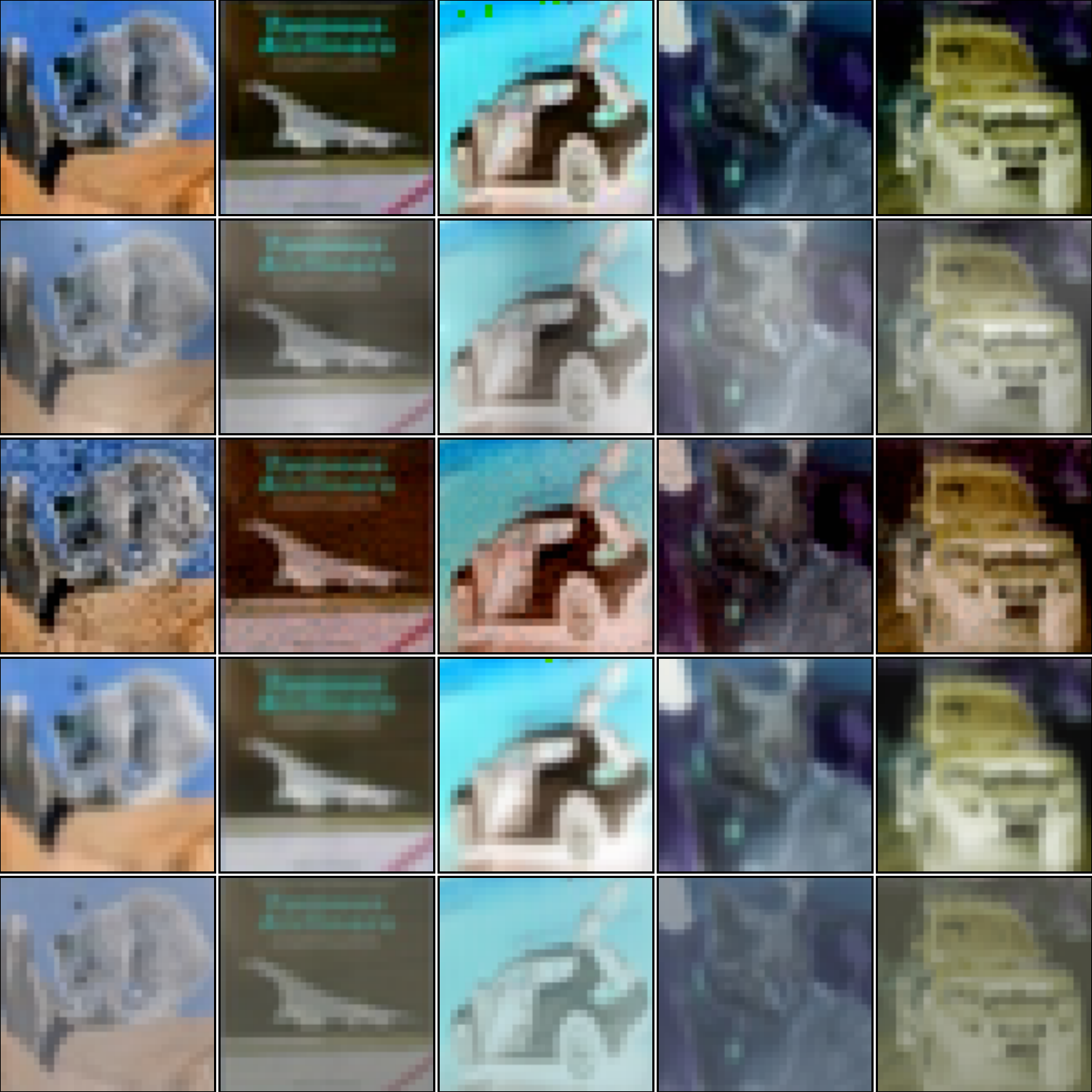}}
    \subfloat[][Digits dataset]{\includegraphics[width=0.48\linewidth,height=4cm]{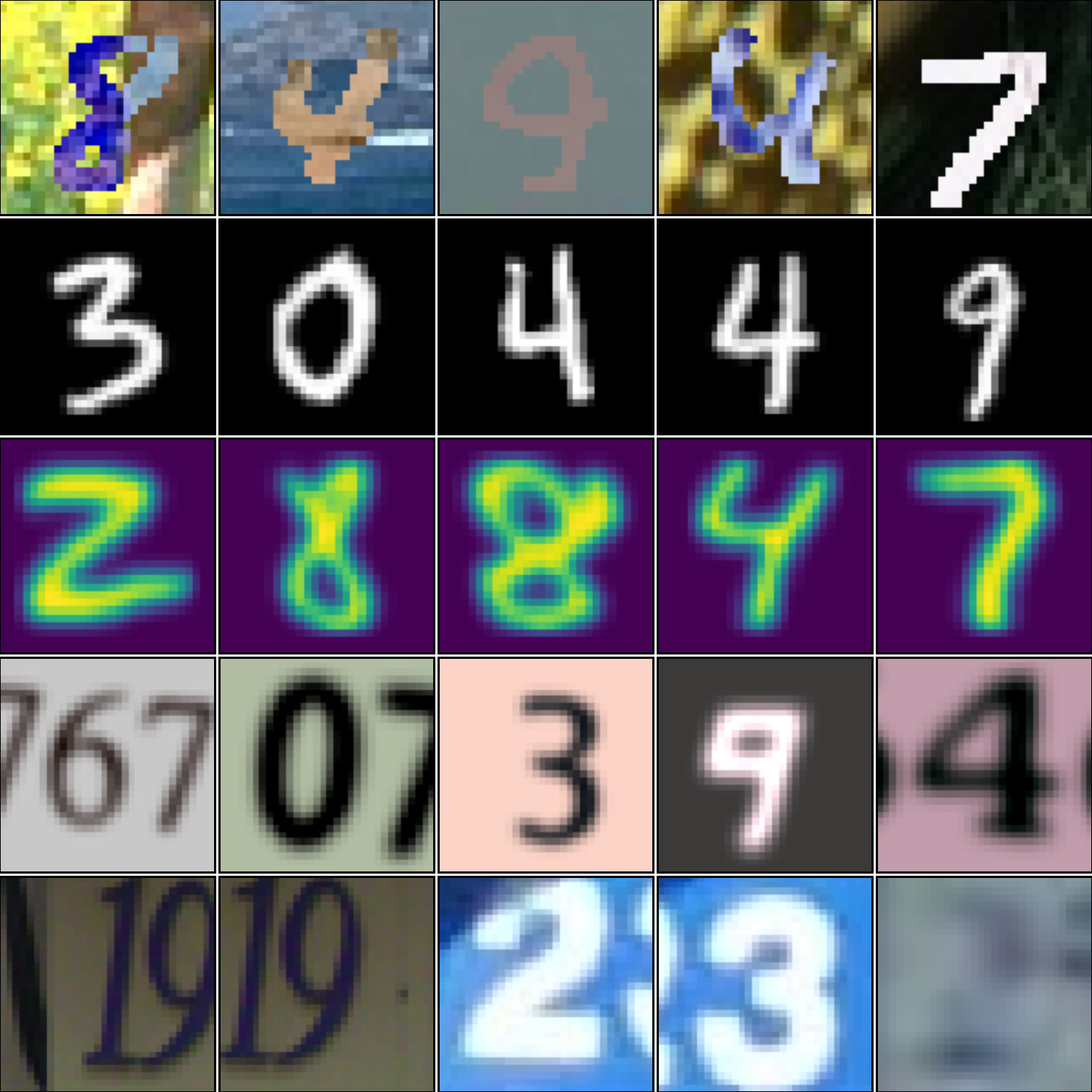}}
    \caption{Examples from two datasets. Each row represents a different domain.}
    \label{dataset}
\end{figure}
\subsection{Setup}
\noindent\textbf{Datasets.} Experiments are performed on the following datasets: Digits, CIFAR-10-Corrupted \cite{DBLP:conf/iclr/HendrycksD19}. Examples are shown in Fig. \ref{dataset}. Specifically, Digits dataset consists of five different domains includes MNIST-M \cite{DANN} (mm), MNIST \cite{mnist} (mt), USPS \cite{usps} (up), SVHN \cite{svhn} (sv), and Synthetic Digits \cite{DANN} (sy). CIFAR-10-Corrupted \cite{DBLP:conf/iclr/HendrycksD19} is composed of 19 different noises as domains, each of which contains five levels of data that indicate noise intensity. Each level has eight classes. We randomly choose five domains, including frost (fro), fog (fo), contrast (co), defocus blur (def), and brightness (bri), of them and evaluate the performance by level 1 and level 3 data. We randomly choose half of the classes for each dataset as the unknown class.

\noindent
\textbf{Baselines.} Since there is no previous setting exactly the same as ours,
we apply the models include single-source methods such as MMD \cite{MMD}, CORAL \cite{CORAL} and DANN \cite{DANN}, multi-source approaches MSDA \cite{moment} and Ltc-MSDA \cite{DBLP:conf/eccv/WangXN020}. They are further combined by open-set methods OSVM \cite{OSVM} and OSBP \cite{OSBP}, as our baselines. 

\noindent\textbf{Evaluation Protocol.} Three metrics are adopted, i.e., normalized accuracy for all classes (\textbf{OS}), normalized accuracy for the known classes only (\textbf{OS$^*$}), and \textbf{H-score}. It is worth noting that H-score is the fairest way to evaluate models' performance, as it balances the OS$^*$ and UNK \cite{DBLP:conf/eccv/BucciLT20}.


\noindent\textbf{Implementation Details.} 
In terms of the model details, the domain encoder consists of one word embedding layer with the dimension of $512$, followed by a linear layer to output the mean $\mu_d$ and standard deviation $\sigma_d$ while the output dimension is set to $30$ and $100$ for Digits and CIFAR-10-Corrupted, respectively. The sample encoder is constructed by three convolutional layers with batch normalization and LeakyReLU and followed by two linear layers to output the latent representation of samples with the dimension of \num{2048} for all datasets. With the same basic residual block as ResNet, the decoder first feeds $z$ into a linear layer, followed by one convolutional layer with ReLU as activation. After parsing the intermediate output into two sets of residual blocks, it goes through a convolutional layer with Sigmoid to produce the reconstructed image. The total number of training epoch $M$ is $50$ for Digits and $100$ for others. The batch size of Digits is set to 32, and 20 for the CIFAR-10-Corrupted dataset. Adam is used as the optimizer with a weight decay of $5e^{-4}$. Following the settings in \cite{DANN}, the learning rate $\mu$ is initiated as $2\times10^{-4}$, decaying as the number of epoch increases. The dropout rate is set to $0.7$. In $\mathcal{L}_{\mathrm{VAE}}$, $\beta$ is set to $6$, and $\xi$ is $1$. The loss coefficient $\lambda$ is set to $2$ and $\gamma$ is $1$, instead, $\alpha$ is empirically defined in a progressive way: $\alpha = \frac{m}{2M}$, where $m$ is the learning step. The threshold $\delta _{known}$ is empirically set to be $0.9$, and $\delta _{unk}$ is $0.3$ in online pseudo labeling.
\begin{table*}[t]
\centering 
\caption{Recognition accuracies on the Digits benchmark.}
\resizebox{1\linewidth}{!}{%
\begin{tabular}{l l c c c c c c c c c c c c c c c}
\toprule
\multirow{2}{*}{Protocol} &\multirow{2}{*}{Method} &\multicolumn{3}{c}{\textbf{mt,up,sv,sy $\rightarrow$ mm}} &\multicolumn{3}{c}{\textbf{mm,up,sv,sym $\rightarrow$ mt}} &\multicolumn{3}{c}{\textbf{mm,mt,sv,sy $\rightarrow$ up}}
&\multicolumn{3}{c}{\textbf{mm,mt,up,sy $\rightarrow$ sv}}
&\multicolumn{3}{c}{\textbf{mm,mt,up,sv $\rightarrow$ sy}}\\ 
\cmidrule(l){3-5}\cmidrule(l){6-8} \cmidrule(l){9-11} \cmidrule(l){12-14} \cmidrule(l){15-17}
& &OS &OS$^*$ &H-score &OS &OS$^*$ &H-score &OS &OS$^*$ &H-score &OS &OS$^*$ &H-score &OS &OS$^*$ &H-score\\
\midrule 
\midrule
\multirow{4}{*}{\rotatebox{90}{Source Comb.}} 
 &OSVM &0.4363	&0.3696	&0.4995 &0.7344	&0.7743	&0.6327 &0.7282	&0.7778	&0.5935 &0.3476	&0.2519	&0.3861 &0.5359	&0.5333	&0.5410\\
 &OSBP\cite{OSBP} &0.4175 &0.3869	&0.4611 &0.7983	&0.8590	&0.6281 &0.8206	&\textbf{0.8564}	&0.7335 &0.3708	&0.3505	&0.4024 &0.4625	&0.4459	&0.4907\\
 &MMD\cite{MMD} + OSVM &0.4534	&0.3879	&0.5184 &0.7349	&0.7686	&0.6452 &0.7222	&0.7691	&0.5968 &0.3722	&0.2992	&0.4257 &0.5529	&0.5610	&0.5357\\
 &CORAL\cite{CORAL} + OSVM &0.4500	&0.3941	&0.5118 &0.7389	&0.7755	&0.6474 &0.7321	&0.7807	&0.6013 &0.3828	&0.3102	&0.4382 &0.5449	&0.5504	&0.5334\\
 &DANN\cite{DANN} + OSVM &0.4617 &0.4313 &0.5066 &0.7328 &0.7642 &0.6568 &0.7257 &0.7778	&0.5823 &0.4687	&0.4756	&0.4479 &0.5544	&0.5590	&0.5449\\
 
\midrule
\multirow{2}{*}{\rotatebox{90}{Multi.}} &MSDA\cite{moment}+OSBP &0.4058	&0.3267	&0.4642 &0.8230	&0.8940	&0.6390 &0.7171	&0.7289	&0.6919 &0.4106	&0.3795	&\textbf{0.4545} &0.3995	&0.3331	&0.4578\\ 
&Ltc-MSDA\cite{DBLP:conf/eccv/WangXN020}+OSBP &0.4321	&0.3307	&0.4184 &0.5902	&0.5643	&0.6326 &0.7206	&0.7317	&0.6967 &0.2598	&0.2288	&0.2949 &0.3745	&0.3408	&0.4187\\
\midrule
\midrule
 &VDD &\textbf{0.5942}	&\textbf{0.6185}	&\textbf{0.5359} &\textbf{0.8490} &\textbf{0.8965}	&\textbf{0.7271}   &\textbf{0.8303}	&0.8415 &\textbf{0.8066} &\textbf{0.4792}	&\textbf{0.5345}	&0.3947 &\textbf{0.5659}	&\textbf{0.5728}	&\textbf{0.5626}\\
\bottomrule
\end{tabular}}
\label{Digits:auc}
\end{table*}
\subsection{Experimental Results}
\begin{figure}[t]
    \centering
    \subfloat[][Reconstructed Images]{\includegraphics[width=0.48\linewidth, height=1.4cm]{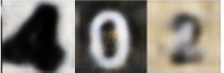}}
    \subfloat[][Fake Reconstructed Images]{\includegraphics[width=0.48\linewidth,height=1.4cm]{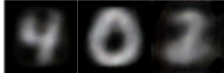}}
    \caption{Examples of reconstructed and fake reconstructed image on Digits dataset}
    \label{fig:recon}
\end{figure}

\textbf{Quantitative Results.} The results of the proposed method and the compared baselines are presented in Tables \ref{Digits:auc} and \ref{Cifar:auc}. 
It can be observed in the tables that the proposed method outperforms most baseline approaches. Taking the Digits dataset as an example, we have the following observations: Combined with the open-set methods, most existing frameworks perform similarly regardless of single- or multi-source domains, indicating that domain shift is not the core issue under the GMDA setting. In terms of the category shift, although OSVM is generally better than OSBP, it is sensitive to the chosen threshold, while there is no similar limitation for OSBP when holding the competitive results. Compared with all baselines, especially for the target USPS dataset, VDD achieves more than $7 \%$ improvement in terms of H-score. The main reason is that VDD can learn domain invariant representations by disentangling multiple source domains. In light of this, the predictions on the target domain can be confidently produced. The experimental outcomes performed on CIFAR-10-Corrupted shows that the proposed method can achieve the best performance in most cases, which further verifies the superiority of the VDD. \\
\textbf{Qualitative Results.}
We provide some samples of the reconstructed and `fake' reconstructed images in Fig. \ref{fig:recon}. After changing the domain embedding, the background style is changed accordingly, while the semantics remain the same.
\begin{table}[t]
\centering 
\caption{Recognition accuracies on the CIFAR-10-Corrupted benchmark. $^{*}$ indicates MSDA methods with OSBP.}
\resizebox{1\linewidth}{!}{%
\begin{tabular}{l l c c c c c }
\toprule
\multirow{2}{*}{Levels} &\multirow{1}{*}{Target Domain} &
\textbf{fro} &\textbf{fo} &\textbf{def}
&\textbf{bri}
&\textbf{co}\\ 
\cmidrule(lr){2-7}
 &Method&H-score  &H-score  &H-score  &H-score  &H-score\\
\midrule 
\midrule
\multirow{3}{*}{\rotatebox{0}{level 1}} 
 &MSDA\cite{moment}$^{*}$ &0.2218 &0.2984 &0.2907&0.3057 &0.2225\\ 
&Ltc-MSDA\cite{DBLP:conf/eccv/WangXN020}$^{*}$ &0.1972 &0.2143&0.2201&0.2295&0.2013\\
\cmidrule(lr){2-7}
 &VDD & \textbf{0.3612}&\textbf{0.3525} &\textbf{0.3479}&\textbf{0.3608}&\textbf{0.3334}\\
\midrule
\midrule
\multirow{3}{*}{\rotatebox{0}{level 3}} 
&MSDA\cite{moment}$^{*}$ &0.2864&\textbf{0.3666}&\textbf{0.4950}&0.2125&\textbf{0.3516}\\ 
&Ltc-MSDA\cite{DBLP:conf/eccv/WangXN020}$^{*}$ &0.2095&0.2194&0.2111&0.2123&0.2006\\
\cmidrule(lr){2-7}
 &VDD &\textbf{0.3000}&0.2956&0.3368&\textbf{0.3823} &0.2743 \\
\bottomrule
\end{tabular}
}
\label{Cifar:auc}
\end{table}

\subsection{Parameter Sensitivity}
\begin{figure}[t]
    \centering
    \setlength{\abovecaptionskip}{0.cm}
    \subfloat[][$\alpha$: coefficient of exemplar loss]{\includegraphics[width=0.52\linewidth, height=3.2cm]{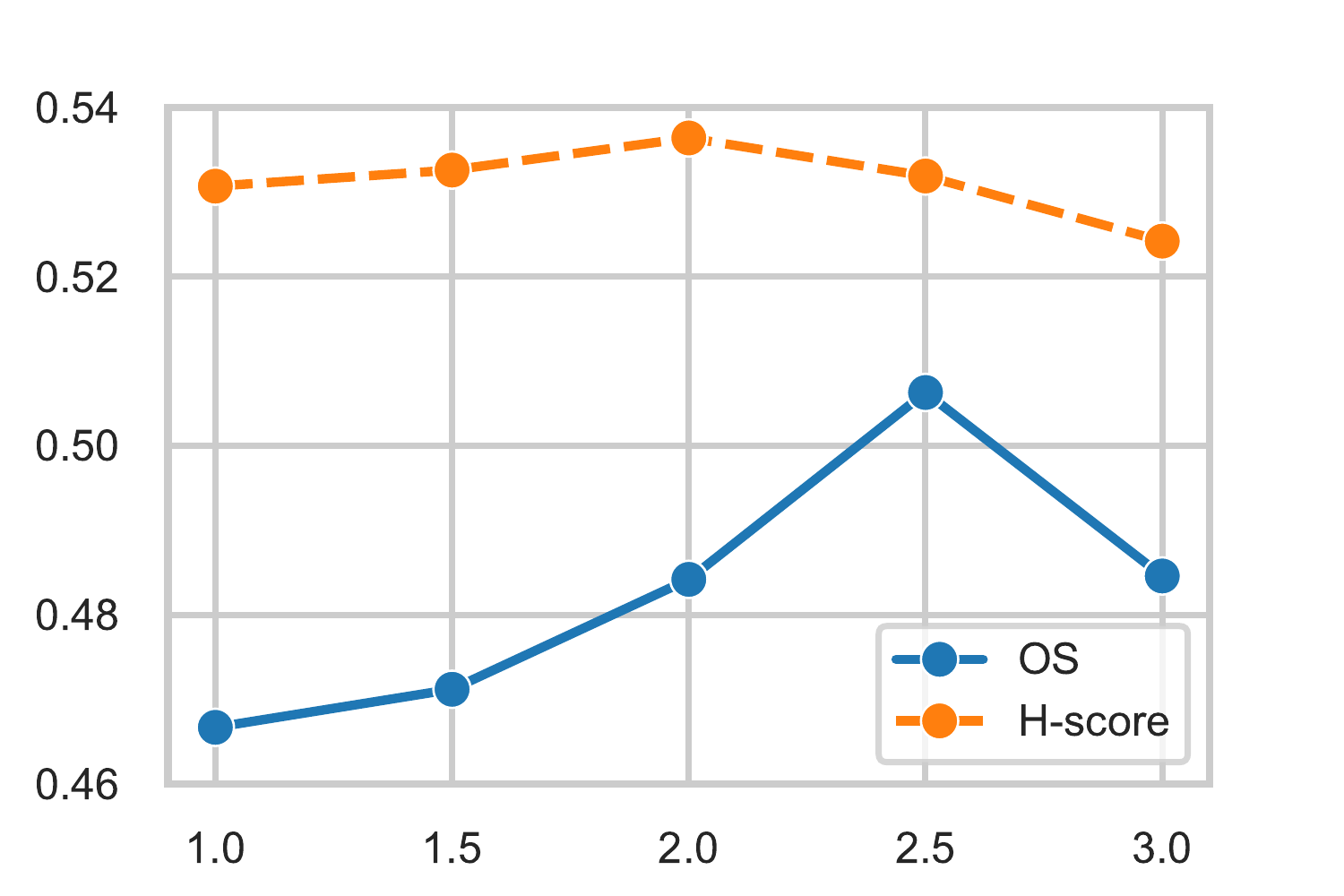}}
    \subfloat[][$\gamma$: coefficient of $\beta$ TC-VAE loss]{\includegraphics[width=0.52\linewidth,height=3.2cm]{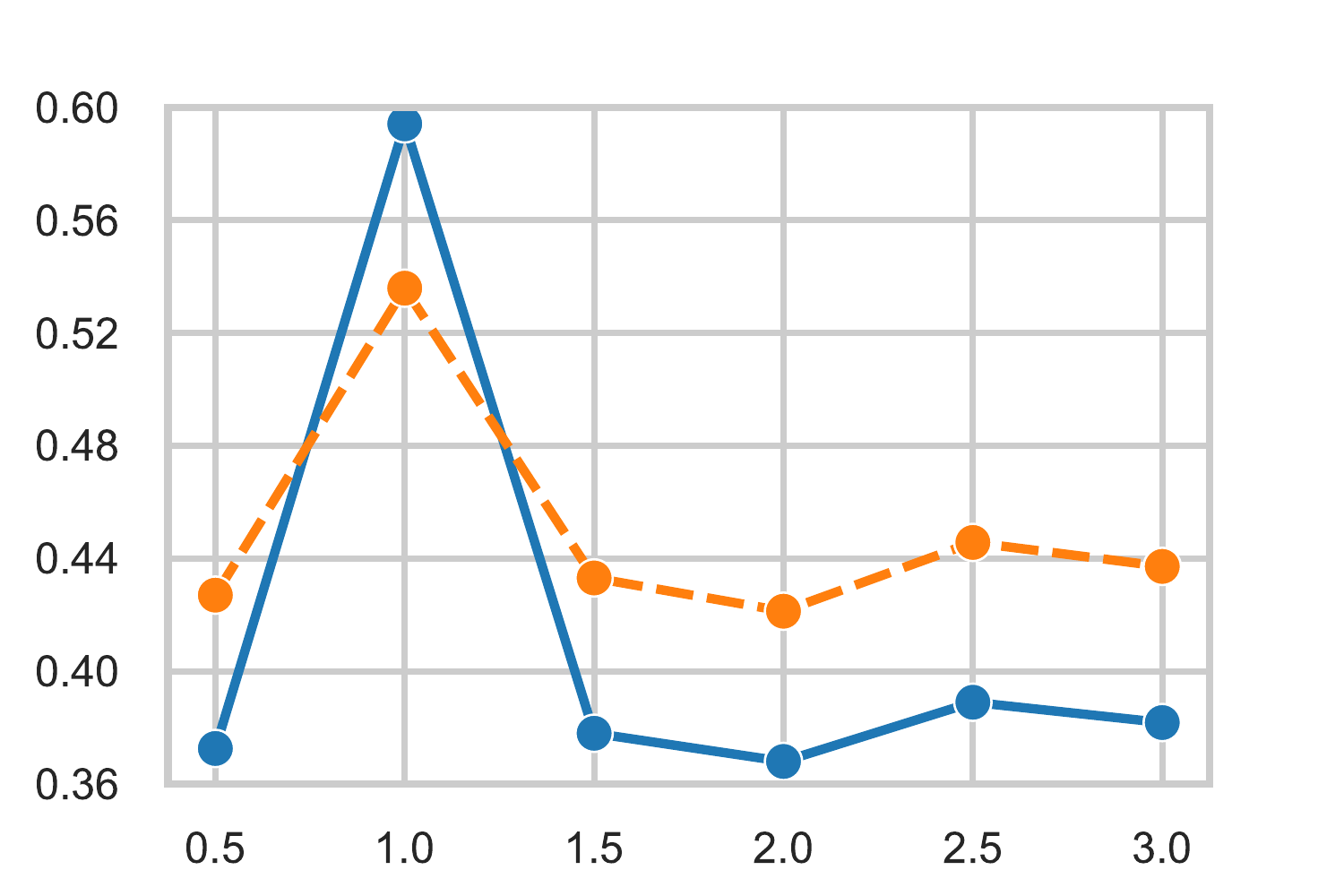}}
    \caption{Parameter sensitivity of loss coefficients $\alpha$ and $\gamma$.}
    \label{fig:para}
\end{figure}
\noindent
We investigated the effect of hyperparameters on the performance of the VDD model, including the loss coefficients $\alpha$ and $\gamma$. $\alpha$ controls the importance of the reconstruction diversity, while $\gamma$ weights the disentanglement process. We conduct two sets of parameter analysis by changing the value of $\alpha$ from $1$ to $3.0$, $\gamma$ from $0.5$ to $3.0$, with the interval of 0.5, the results of which are plotted in Fig. \ref{fig:para}. It shows that in $[0.5, 2.5]$, the OS score of the proposed method grows as $\alpha$ increases. The H-score peak at about $0.536$ where $\alpha$ is set to 2. While OS achieves the best when $\alpha$ is $2.5$. These phenomena indicate that the exemplar loss applied contributes positively to the multi-source learning, yet amplifying the loss too much would hinder the classification learning. Compared with the results shown in Table \ref{Digits:auc}, we can see that the best result comes from the progressively defined $\alpha$, which implies dynamically adjusting the hyperparameter $\alpha$ could adapt the learning of the objective better. The result on $\gamma$ exhibits a similar tendency, demonstrating that balancing the importance of different components can yield better results. Furthermore, it exemplifies that when $\gamma$ is $1$, the performance is the best, as evidenced by OS and H-score consistently. At the best result achieved by adjusting $\gamma$, we discover that the global tendency of OS and H-score are almost consistent under the influence of different values of $\alpha$ and $\gamma$, It indicates the influence of disentanglement significance on the model is integral, impacted the classification performance of each domain simultaneously. 
\begin{table}[t]
\caption{The ablation performance of the proposed VDD on the Digits dataset.}
\resizebox{1\linewidth}{!}{%
\begin{tabular}{l c c c c} 
\toprule 
& \multicolumn{2}{c}{\textbf{mt,up,sv,sy$\rightarrow$mm}} & \multicolumn{2}{c}{\textbf{mm,mt,up,sv$\rightarrow$sy}}\\ 
\cmidrule(l){2-3}\cmidrule(l){4-5}
Method &OS &H-score &OS &H-score\\ 
\midrule 
GMDA w/o $\mathcal{L}_{exe}$ &0.3767&0.4305 &0.3463&0.3893\\
GMDA w/o disent &0.3920&0.4487&0.3246&0.3675\\
\midrule
\midrule
GMDA&\textbf{0.5942}&\textbf{0.5359}&\textbf{0.5659}&\textbf{0.5626}\\
\bottomrule
\end{tabular}
}
\label{tab:ablation}
\end{table}
\subsection{Ablation Study}
\noindent
By conducting an ablation study that evaluates variants of VDD, we dive deep into the effectiveness of the proposed domain disentanglement framework. (1) \textbf{GMDA w/o $\mathcal{L}_{exe}$} is the variant without the diversified reconstruction by fake domain label in Eq. \eqref{loss:diverse}; (2) \textbf{GMDA w/o disent} is the variant without VAE loss in Eq. \eqref{loss:vae}. The experimental results are reported in Table \ref{tab:ablation}, where it can be observed that the absence of any part could lead to performance reduction. The missing of reconstruction diversity (i.e., $\mathcal{L}_{exe}$) causes about $0.219$ decrease of OS and $0.136$ of H-score on target domain MNIST-M and Synthetic Digits while lacking disentanglement results in $0.222$ and $0.137$ reduction for OS and H-score respectively. In addition, the degradation for both GMDA w/o $\mathcal{L}_{exe}$ and GMDA w/o disent are almost the same, meaning that diversifying the reconstruction with exemplar learning is as crucial as disentangling the feature in the latent space.

\section{Conclusion}
In this paper, we propose a Variational Domain Disentanglement framework, which aims to address the domain and category shift in a novel GMDA setting. We verify that the alignment problem of multiple domains could be solved by disentangling the domain feature in latent space. Specifically, we record that previous state-of-the-art can be beaten by our domain disentanglement model. The ablation study shows that both reconstruction diversity and disentanglement method play great significant roles in the final result. We will explore the more challenging and practical domain adaption problems in the future.

\bibliographystyle{IEEEbib}\small

\bibliography{icme2021template}

\end{document}